\documentclass[conference]{IEEEtran}
\IEEEoverridecommandlockouts
\usepackage{cite}
\usepackage{amsmath,amssymb,amsfonts}
\usepackage{algorithmic}
\usepackage{graphicx}
\usepackage{epstopdf}
\usepackage{textcomp}
\usepackage{booktabs}
\usepackage{xcolor}
\usepackage[linesnumbered,commentsnumbered,ruled,vlined]{algorithm2e}
\def\BibTeX{{\rm B\kern-.05em{\sc i\kern-.025em b}\kern-.08em
    T\kern-.1667em\lower.7ex\hbox{E}\kern-.125emX}}
\begin{document}

\title{Encoded Spatial Attribute in Multi-Tier Federated Learning
}

\author{
    \IEEEauthorblockN{
    Asfia Kawnine\IEEEauthorrefmark{1},
    Francis Palma\IEEEauthorrefmark{2},
    Seyed Alireza Rahimi Azghadi\IEEEauthorrefmark{1},
    Hung Cao\IEEEauthorrefmark{1}
    }

    \IEEEauthorblockA{
     \IEEEauthorrefmark{1} \textit{Analytics Everywhere Lab, Faculty of Computer Science, University of New Brunswick, Canada} \\
     \IEEEauthorrefmark{2} \textit{SE+AI Research Lab, Faculty of Computer Science, University of New Brunswick, Canada} \\
    }
}

\maketitle

\begin{abstract}
This research presents an Encoded Spatial Multi-Tier Federated Learning approach for a comprehensive evaluation of aggregated models for geospatial data. In the client tier, encoding spatial information is introduced to better predict the target outcome. The research aims to assess the performance of these models across diverse datasets and spatial attributes, highlighting variations in predictive accuracy. Using evaluation metrics such as accuracy, our research reveals insights into the complexities of spatial granularity and the challenges of capturing underlying patterns in the data. We extended the scope of federated learning (FL) by having multi-tier along with the functionality of encoding spatial attributes. Our N-tier FL approach used encoded spatial data to aggregate in different tiers. We obtained multiple models that predicted the different granularities of spatial data. Our findings underscore the need for further research to improve predictive accuracy and model generalization, with potential avenues including incorporating additional features, refining model architectures, and exploring alternative modeling approaches. Our experiments have several tiers representing different levels of spatial aspects. We obtained accuracy of 75.62\% and 89.52\% for the global model without having to train the model using the data constituted with the designated tier. The research also highlights the importance of the proposed approach in real-time applications.
\end{abstract}


\section{Introduction}
\label{intro}
Federated learning (FL) is an approach that contrasts traditional centralized learning, where data from all devices are aggregated on a central server for model training \cite{banabilah2022federated}. 
This approach brings notable benefits, including enhanced privacy, the ability to train models locally, and improved adaptability to both data and network variability \cite{li2020review, zhang2021survey, nilsson2018performance}.
Despite these advancements, most research has focused on centralized models rather than localized ones, and FL has received limited attention in the context of spatial data \cite{graser2022role}.

Previous studies have focused on training models using FL with balanced datasets and building models tailored to specific applications involving spatial data \cite{nguyen2021spatially}. 
However, obtaining balanced data from all clients is often not feasible in real-world applications. 
Some researchers have incorporated statistical methods into FL to assess spatial attributes \cite{belal2024survey}.
Nevertheless, these statistical methods, particularly those used for spatiotemporal forecasting, often fail to account for changes in trends over time. 
Research on hierarchical federated learning has explored aspects such as computational efficiency, privacy concerns, and network convergence \cite{ooi2023measurement}. 
FL introduced hierarchical aggregation to enhance processing efficiency, which organizes aggregations into clusters within a hierarchical structure, enabling resource-efficient computing \cite{wang2021resource}. 
In response to the challenges of network convergence, there has been a growing interest in edge computing and data-driven approaches to address these complexities \cite{9148862}.
Among these, the integration of spatial data has received little attention because of its potential to provide granular insights into localized prediction models, spatial distribution, and underlying drivers.
We can discern spatial dependencies and predict targeted values by harnessing encoded spatial data and a multi-tier approach, encapsulating geographic information. 
Concurrently, having different distributions of datasets for analyzing the predictive modeling would portray the possibility of industrial use of this approach. 
Thus, our research seeks to leverage these approaches to build localized models and contribute to efforts toward spatial analysis.

The application of FL to spatial data presents different challenges and opportunities. 
Because of the distributed architecture, it can have different spatial attributes for different clients. 
However, having trained models unique to the spatial attribute is challenging. 
Extending FL techniques to handle unique models can unlock new potential for predictive analysis and decision-making.
The objectives of this research are multifaceted. It aims to make predictions through the integration of spatial data and federated learning techniques.
Additionally, we aim to assess the effectiveness of multi-tier federated learning architecture in aggregating insights from diverse spatial data.
Our hypothesis posits that integrating encoded spatial data and multi-tier federated learning methodology will facilitate the prediction of the trained models.

In contrast, the multi-tier federated learning architecture will effectively aggregate insights across geographical scales and will be able to provide localized models.
In contemporary discourse surrounding spatial analysis, predicting target value is pivotal. 
This can guide efforts to make informed decisions in various sectors for real-time applications.
Within this context, our research addresses the intricate challenge of understanding patterns through the lens of spatial data and machine learning methodologies.
Our proposed multi-tier approach for federated learning and encoded spatial data would handle the unique spatial granularity.
We aim to focus on spatial data in different tiers of the FL and have localized models depending on the spatial attribute. 

To test our approach, we used two different datasets; the first is green-house gas emission data with provincial \cite{greenhouseGas2022} and central Canada data, and the other is the New Brunswick (NB) power dataset containing the recharge history of the electric vehicles around NB \cite{richard2021spatial}. 
At its core, our research endeavors to unravel the spatial intricacies of both datasets.
Emissions distribution, particularly on the regional trends present within provinces or territories.
And usage of energy by electric vehicles around NB.
Through our exploration, we tried to see how accurately the model can predict the values.


\section{Related Work}
\label{rel_work}

This section reviews studies and recent advancements in federated learning (FL), focusing on methodologies employed with spatial data.

\subsection{Federated Learning (FL)}

FL has gained prominence for its efficiency in communication-sensitive settings. Unlike hierarchical clustering or gradient sparsification, FL benefits from local model updates \cite{shahid2021communication}. Methods such as Federated Averaging (FedAvg) are used in areas like smart city sensing \cite{s20216230}, effectively managing non-IID data and system heterogeneity when combined with Stochastic Gradient Descent (SGD) \cite{li2019convergence}. Strategies like client clustering and edge device adaptation further address non-IID challenges \cite{zhu2020robust}, though handling uneven real-world data, such as geospatial data, remains challenging.
The Weighted Federated Average (FedAvg Weighted) approach helps manage imbalanced non-IID data but may bias the global model \cite{hong2022weighted}. FL also shows improved performance in wireless IoT networks by reducing communication costs compared to traditional models \cite{tran2019federated}. Integrating FL with edge computing provides a cost-effective solution for real-time monitoring, addressing privacy and latency issues efficiently \cite{abimannan2023towards}. Edge-cloud hierarchical FL enhances training efficiency using FedAvg in a two-tier framework, balancing communication and computation, though it may not always offer localized models for individual clients \cite{liu2020client}.

\subsection{Spatial data employed with FL}

Recent advancements in FL for spatial data analysis have been notable. Spatial data, crucial for environmental monitoring, urban planning, and transportation, traditionally relies on centralized data collection, raising concerns about privacy and data transfer. FL provides a decentralized solution, allowing data to remain on local devices and only sharing model updates, thus addressing privacy issues and reducing data transmission needs \cite{majcherczyk2021flow}. Studies have explored FL for privacy-preserving geospatial data analysis \cite{saxena2023advancing}, though challenges persist in training region-specific models versus centralizing aggregated data. For example, FL experiments have focused on collaborating among mobile users for location-based services \cite{yin2020fedloc}, highlighting the need for further refinement.
Recent progress also includes applying neural networks and ensemble learning to spatial data, enhancing accuracy and efficiency in managing complex spatial information. Deep learning models, particularly, are effective for tasks such as land cover classification, traffic prediction, and environmental monitoring \cite{sun2020research}. However, challenges like overfitting due to data variability and sparsity remain.

To address challenges in spatial data analysis, ensemble learning methods have been combined with neural networks to improve model robustness and generalization. Ensemble learning aggregates multiple models to enhance predictive accuracy, reducing variance and minimizing overfitting \cite{rokach2010ensemble}. Techniques such as Random Forests, Gradient Boosting Machines, and Bagging have been integrated with neural networks to develop more reliable models.
This hybrid approach effectively captures diverse spatial patterns and lowers error rates, with successful applications in predicting urban expansion, assessing environmental risks, and optimizing transportation systems \cite{dong2018stacking}. However, issues like computational complexity and model interpretability persist. Ongoing research focuses on enhancing the scalability and transparency of these ensemble-neural network models for spatial data applications.

Our review has revealed several key gaps that our research will address. Firstly, many spatial data prediction studies rely on centralized data processing, which needs more robustness for real-time applications. We will utilize federated learning (FL) and edge computing to enable decentralized, privacy-preserving model training with reduced latency. Secondly, while FL has been explored for spatial data, there is a need for evaluation across various spatial granularities. Our research will address this by rigorously assessing a multi-tier FL approach with spatial encoding in real-world scenarios, advancing robust prediction systems.

In summary, integrating spatial data analysis, multi-tier FL, and spatial encoding presents a promising path to extending FL methods and prediction systems. This would bridge the gap between theoretical research and practical applications while more effectively tackling complex challenges.
\section{Proposed Methodology}
\label{method}


In this research, we introduce a novel N-tier federated learning (FL) approach designed for making spatial data predictions. The proposed approach structures the FL system into multiple tiers, with encoding applied at the client tier to enhance the data processing and learning capabilities. For the experimental validation, we employed two datasets: the Greenhouse Gas Emissions dataset \cite{greenhouseGas2022} and the NB Power Energy Consumption dataset \cite{richard2020discovering}. These datasets were chosen due to their relevance in assessing the performance of our model in handling environmental and energy-related data.

\begin{figure*}[htbp]
    \centering
    \includegraphics[width=0.95\textwidth]{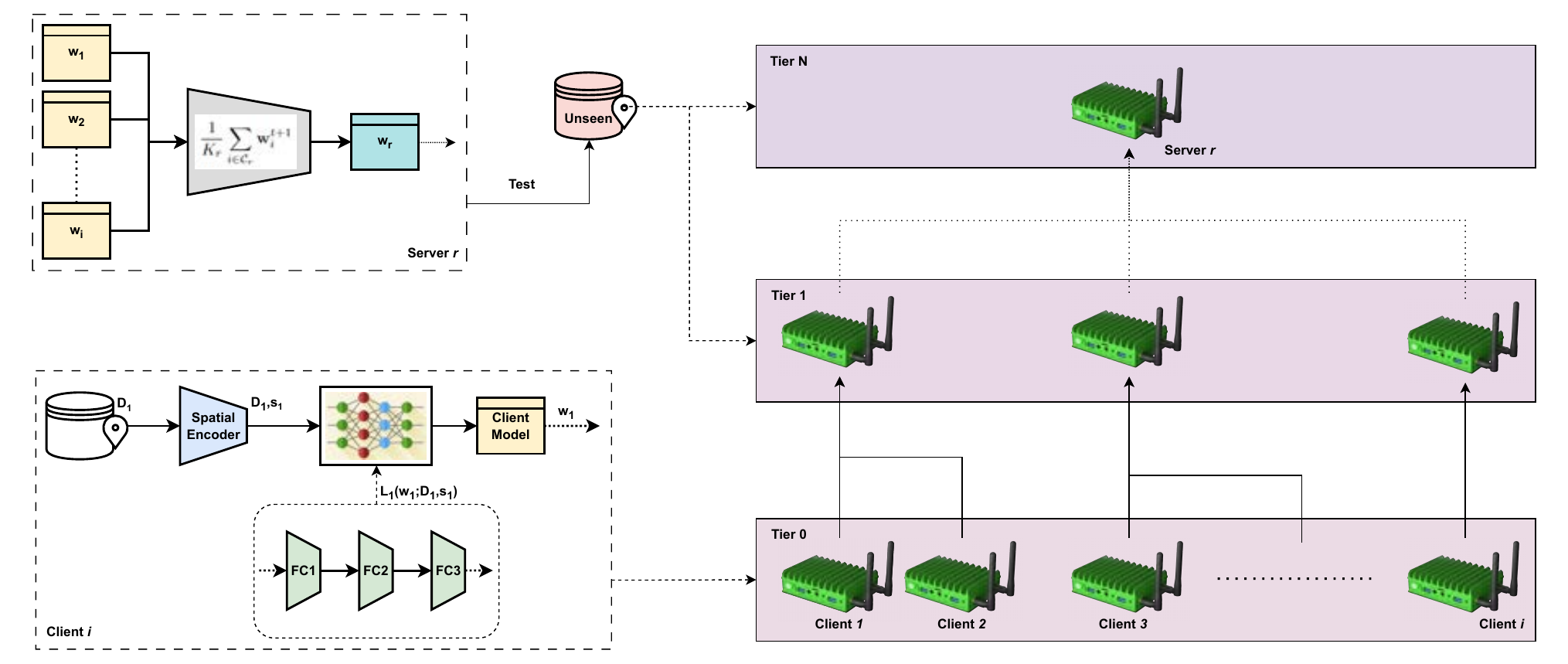}
    \caption{N-tier FL architecture with spatial encoding.}
    \label{fig:methodology}
\end{figure*}

We present an overview of the proposed n-tier federated learning (FL) methodology. The system is designed to make predictions using spatial data, and different tiers represent distinct levels of spatial granularity. Figure \ref{fig:methodology} illustrates the overall architecture of the system.

At the client tier, spatial data is collected and processed using spatial encoding techniques to preserve key spatial patterns. Each client functions independently and utilizes models, such as neural networks, to make localized predictions. For instance, in the greenhouse gas emissions dataset, individual provinces or territories act as clients, while in the NB power dataset, stations represent the clients. After local processing, the model updates (e.g., weights and biases) are transmitted to higher tiers for aggregation.

Moving up to tier-1, the local insights are aggregated using the Federated Averaging (FedAvg) technique \cite{sun2021decentralized}, which gathers encoded spatial patterns from individual clients. Depending on the dataset, this aggregation facilitates the discovery of broader spatial trends at the regional or city level. Geo-location information is crucial in this step, allowing the system to capture spatial divisions during aggregation.

At tier-2, the highest tier, insights from various regions or cities are synthesized to provide a comprehensive prediction at a global or national level. In the case of the greenhouse gas emissions dataset, this tier offers provincial insights, while the NB power dataset provides national-level predictions. Integrating multi-tier aggregation ensures that the methodology accounts for spatial patterns at multiple levels, progressively refining the predictions.

The proposed approach is systematic, allowing localized predictions to be improved and combined at each tier to build a broader, more detailed understanding of the data as it moves through the hierarchy.

\subsection{Local Model Updates}
Each client \(i\) trains its local model with spatial encoding over \(E\) epochs, updating its model weights according to the following equation:
\begin{equation}
    \mathbf{w}_i^{t+1} = \mathbf{w}_i^t - \eta \nabla \mathcal{L}(\mathbf{w}_i^t; \mathcal{D}_i, \mathbf{s}_i)
\end{equation}
where:
\begin{itemize}
    \item \(\mathcal{D}_i\) represents the local dataset of client \(i\),
    \item \(\mathcal{L}\) is the loss function,
    \item \(\mathbf{w}_i\) refers to the local model weights at client \(i\),
    \item \(\mathbf{s}_i\) is the spatial encoding for client \(i\)'s data.
\end{itemize}

\subsection{Tier Aggregation}
At higher tiers, such as tier-1 and tier-2, the local models are aggregated using the Federated Averaging (FedAvg) technique:
\begin{equation}
    \mathbf{w}_r^{t+1} = \frac{1}{K_r} \sum_{i \in \mathcal{C}_r} \mathbf{w}_i^{t+1}
\end{equation}
where:
\begin{itemize}
    \item \(K_r\) is the number of clients in region \(r\),
    \item \(\mathcal{C}_r\) is the set of clients in region \(r\).
\end{itemize}

This multi-tier FL approach, enriched with spatial encoding, ensures that local models capture critical spatial relationships in the data, which are then progressively aggregated to provide regional and global predictions.

Our research's data collection procedures were designed to obtain comprehensive datasets suitable for the N-tier Federated Learning (FL) approach. We sourced the greenhouse gas emissions dataset from Statistics Canada's Physical Flow Account, which covers 2009 to 2021. This dataset provides a detailed breakdown of emissions at various levels, including Canada-wide, province, and territory tiers, serving as the basis for our spatial data analysis.

Following the acquisition of the dataset, we performed meticulous data structuring to ensure relevance and reliability. Preprocessing steps included removing extraneous information to streamline the dataset for efficient analysis. Additionally, we organized the spatial data into individual client datasets, with each province or territory treated as a separate entity in alignment with our federated learning approach.

We also employed the NB power dataset \cite{kawnine2024evaluating}, which contains electric vehicle charging event data from April 2019 to June 2022, provided by New Brunswick Power (NB Power). For this dataset, we arranged the data in a hierarchy based on station, city, and province, allowing us to effectively implement the dataset within our proposed FL architecture.

In summary, this research leverages an N-tier federated learning approach with encoded spatial data, systematically scaling from local to higher tiers. At the local tier, neural network models are employed, and the Federated Averaging technique is utilized at subsequent tiers. This approach provides a robust methodology for analyzing and interpreting predictions within the context of spatial data, allowing for more precise and contextually aware predictions.

\section{Experiment}
\label{exp}

Our experimental design aimed to validate the proposed N-tier federated learning (FL) methodology using two spatial datasets: (1) the greenhouse gas emissions dataset across Canada and (2) New Brunswick power consumption data for electric vehicles. Key variables included geographical locations, reference dates, and other relevant factors. These variables were essential for predicting the target values in each dataset.

The data was distributed based on locations, creating individual clients for each region. Following data collection, preprocessing steps were applied to clean the data, remove noise and outliers, and fill in missing values using methods like interpolation. This ensured the integrity and consistency of the dataset before training.

Using a three-tier system to train and aggregate models, we employed reComputer J1010 machines as both client and server devices. Simple two-layer neural networks were deployed on the client devices, using libraries such as \texttt{pandas} and \texttt{torch} in Python. For greenhouse gas emissions, 13 clients were set up based on the provinces and territories of Canada, while nine clients were used for the New Brunswick power consumption data. The Adam optimizer \cite{kingma2014adam} was used for model training, with hyperparameters fine-tuned using cross-validation. Input features included geo-location, reference dates, target values, and other relevant factors.

The proposed multi-tier FL approach was implemented, where models aggregate local predictions based on spatial data encoding at each tier. As shown in Algorithm \ref{algo:efl}, each client \(i\) trains a local model by encoding its spatial data, and these models are aggregated at higher tiers using the FedAvg algorithm. The second and third-tier servers further aggregate these spatial-encoded models for broader regional and global predictions.

\begin{algorithm}[t]
\SetAlgoLined
\SetKwInOut{Input}{Input}
\SetKwInOut{Output}{Output}
\SetKwComment{comment}{\# }{}
\Input{Num\_Clients, Num\_Tiers}
\Output{Predictive models}
 initialization\; input\_dimention, output\_dimention \\
 \While{Num\_Tiers}{
  \eIf{Num\_Tiers$=$0}{  
   \comment{0 indicates the first tier}
   encode spatial data\;
   train NN model\;
   calculate spatial weight\;
   send(model\_bias and spatial weight) to assigned Tier\;
   }{
   \While{model\_bias and spatial weight}{
    receive model\_bias and spatial weight\;
    model += spatial weight * model\_bias\;
    \comment{it aggregates the spatial encoded models using FedAvg}
   }
  }
 }
 \caption{Spatial Encoded N-tier FL}

\label{algo:efl}
\end{algorithm}

In Algorithm \ref{algo:efl}, each client first encodes its spatial data and trains a local neural network model. The resulting model parameters (bias) and spatial weights are then sent to the next tier for aggregation. At each higher tier, the models are aggregated using the FedAvg technique, where models are combined with their spatial weights to capture regional or global trends.

For comparison, we implemented traditional neural networks, ensemble learning \cite{dong2020survey}, FedAvg, and FedAvg weighted \cite{li2023revisiting}. Model performance was evaluated based on prediction accuracy using a validation dataset to ensure consistency across models. Hyperparameters were adjusted iteratively to optimize performance.

\section{Results and Discussion}
\label{results}

In this study, we proposed an approach that aggregates models trained on geo-location data to enhance the accuracy of spatial predictions.

\vspace{2mm}
\noindent\textbf{Gas Emission Data (Experiment 1)}: In the first experiment, we evaluated several predictive models using a gas emission dataset, where the predicted values were categorized as 0 (low), 1 (medium), and 2 (high). Table \ref{tab:results_gmg} presents the accuracy across four servers. The {Neural Network} achieved the highest accuracy on all servers, ranging from 73.70\% to 81.62\%. This is because it was trained on regional datasets. The {Ensemble Learning} method had the lowest accuracy across all servers, with values between 56.90\% and 79.21\%.

Our proposed {N-tier encoded federated learning} performed competitively, achieving an accuracy of 62.53\% in Server 1 and 70.55\% in Server 4, showing better performance on most servers compared to {FedAvg} and {FedAvg Weighted}.

\begin{table}[t]
\caption{Comparison of accuracy: Gas emission data}
\centering
\setlength{\tabcolsep}{2.5pt}
\centering
\begin{scriptsize}
\begin{tabular}{cccccc}
\toprule
Tier-1   & Neural Net & Ensemble & FedAvg & FedAvg Weighted & N-tier encoded FL \\ \midrule
Server-1 & 73.78          & 56.90               & 62.90       & 57.25           & 62.53         \\
Server-2 & 81.62          & 79.21               & 66.15       & 80.00           & 80.42         \\
Server-3 & 73.70          & 58.41               & 64.57       & 68.00           & 68.14         \\
Server-4 & 79.57          & 60.65               & 66.63       & 70.00           & 70.55         \\ 
 \bottomrule
\end{tabular}%
\end{scriptsize}
\vspace{0.2cm}
\label{tab:results_gmg}
\end{table}

\begin{figure}[t]
\centering
\includegraphics[width=9cm]{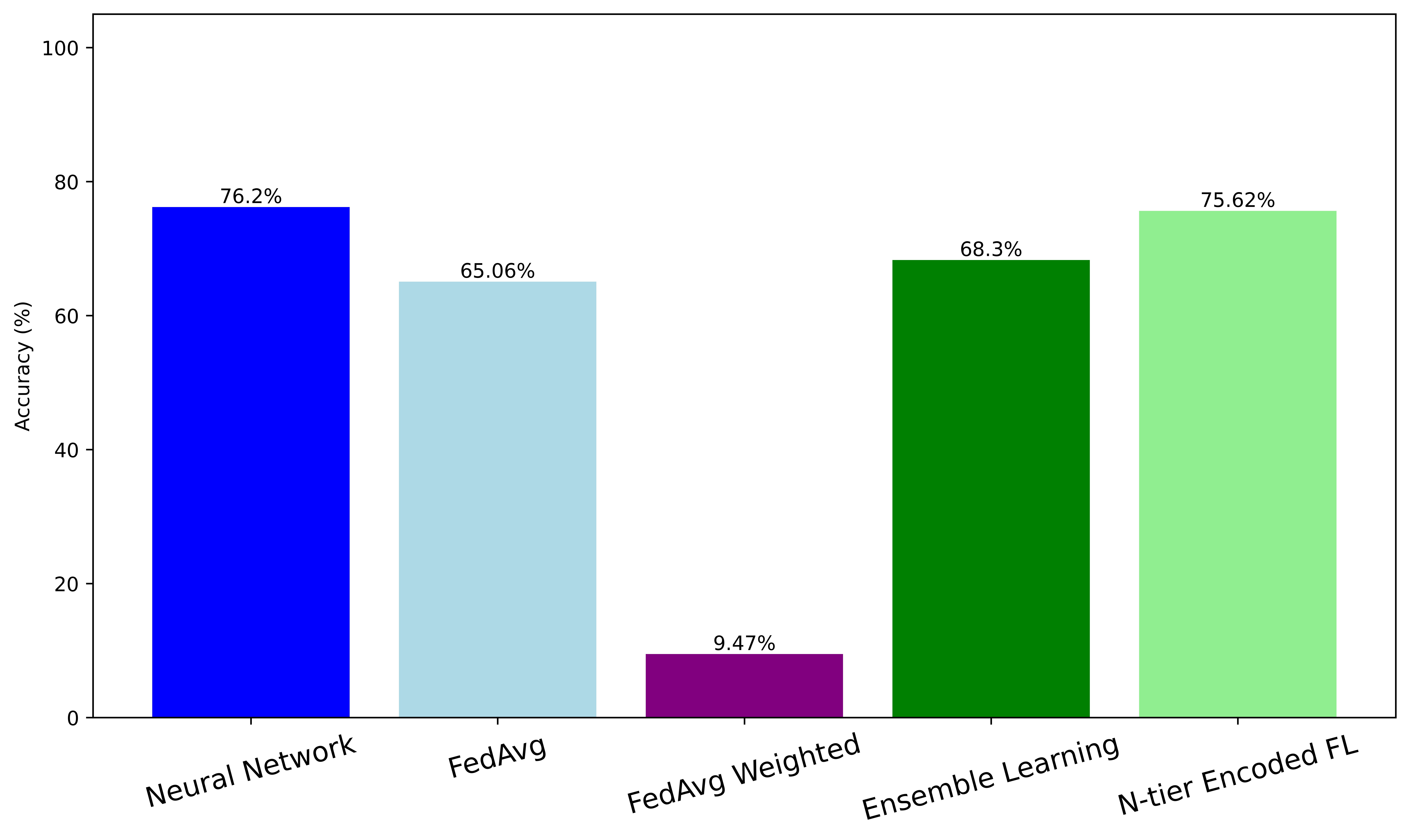}
\caption{Comparison of accuracy for gas emission data across tier-2}
\label{fig:t2_g}
\end{figure}

Figure \ref{fig:t2_g} visually represents the accuracy comparison for tier-2 predictions in the gas emission dataset. As shown, the {Neural Network} achieved the highest accuracy (76.20\%), while the {FedAvg Weighted} method had the lowest (9.47\%). Our proposed {N-tier encoded approach} achieved 75.62\%, coming close to the Neural Network. The significant drop in accuracy for FedAvg Weighted was due to an imbalanced dataset, which affected generalization capabilities.

Table \ref{tab:acp_gmg} summarizes the localized model predictions for gas emission across the clients. The predicted and actual values are reported for each client and categorized into three emission levels (low, medium, and high).

\begin{table}[t]
\caption{Localized model predictions: Gas emission}
\centering
\begin{small}
\begin{tabular}{ccc}
\toprule
\textbf{Client ID} & \textbf{Predicted Classes} & \textbf{Actual Classes} \\ \midrule
0               & {[}0, 1, 1, 0, 0{]}       & {[}0, 1, 1, 0, 0{]}    \\ 
1               & {[}0, 0, 0, 0, 0{]}       & {[}0, 0, 0, 0, 0{]}    \\ 
2               & {[}1, 1, 1, 1, 1{]}       & {[}1, 1, 1, 1, 0{]}    \\ 
3               & {[}1, 0, 1, 0, 1{]}       & {[}1, 0, 1, 0, 1{]}    \\ 
4               & {[}1, 1, 0, 1, 0{]}       & {[}1, 1, 0, 1, 0{]}    \\ 
5               & {[}1, 1, 0, 2, 1{]}       & {[}1, 1, 0, 2, 1{]}    \\ 
6               & {[}0, 1, 0, 1, 1{]}       & {[}0, 0, 0, 1, 1{]}    \\ 
7               & {[}1, 1, 0, 1, 1{]}       & {[}1, 1, 0, 1, 1{]}    \\
8               & {[}1, 1, 1, 2, 1{]}       & {[}1, 0, 1, 0, 1{]}    \\ 
9               & {[}1, 0, 0, 1, 1{]}       & {[}1, 0, 0, 1, 1{]}    \\ 
10              & {[}0, 0, 1, 0, 0{]}       & {[}0, 0, 1, 0, 0{]}    \\ 
11              & {[}0, 1, 0, 0, 1{]}       & {[}0, 1, 0, 0, 0{]}    \\
12              & {[}0, 0, 1, 1, 0{]}       & {[}0, 0, 1, 1, 0{]}    \\
 \bottomrule
\end{tabular}%
\end{small}
\vspace{0.2cm}
\label{tab:acp_gmg}
\end{table}

\vspace{2mm}
\noindent\textbf{Energy Consumption Data (Experiment 2)}: In the second experiment, we evaluated the models on an energy consumption dataset, with predicted values of 0 (low) and 1 (moderate). Table \ref{tab:results_ener} shows the accuracy across three servers. The {Neural Network} once again had the highest accuracy in all servers, with values ranging from 96.89\% to 98.77\%. This method was trained with provincial datasets, contributing to its superior performance.
Our approach, {N-tier encoded federated learning}, achieved an accuracy close to the Neural Network. 
It outperformed {Ensemble Learning}, {FedAvg}, and {FedAvg Weighted}.

\begin{table}[t]
\caption{Comparison of accuracy: Energy consumption data}
\centering

\setlength{\tabcolsep}{3pt}
\centering
\begin{scriptsize}
\begin{tabular}{cccccc}
\toprule
Tier-1   & Neural Net & Ensemble & FedAvg & FedAvg Weighted & N-tier encoded FL \\
\midrule
Server-1 & 99.88 & 83.97 & 89.88 & 78.61 & 89.88 \\
Server-2 & 96.89 & 81.44 & 86.99 & 77.89 & 89.39 \\
Server-3 & 98.77 & 76.71 & 74.93 & 76.54 & 89.29 \\ 
\bottomrule
\end{tabular}%
\end{scriptsize}

\vspace{0.2cm}
\label{tab:results_ener}
\end{table}

\begin{figure}[t]
\centering
\includegraphics[width=9cm]{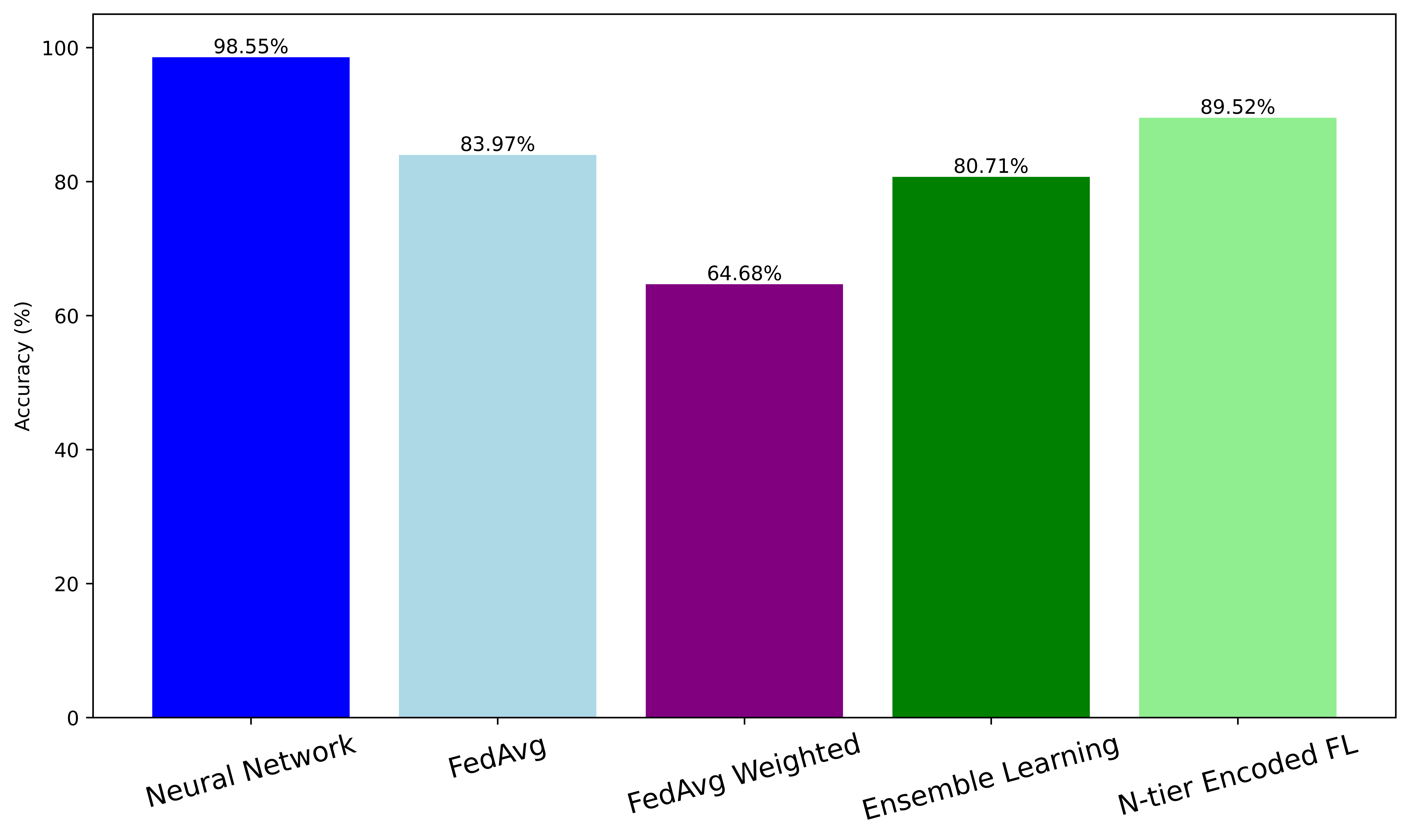}
\caption{Comparison of accuracy for energy consumption models in tier-2}
\label{fig:t2_e}
\end{figure}

Figure \ref{fig:t2_e} visually compares model accuracy in tier-2 for energy consumption predictions. Similar to the gas emission results, the {Neural Network} had the highest accuracy (98.55\%), while {FedAvg Weighted} performed the worst (64.68\%). Our approach, {N-tier encoded federated learning}, performed competitively, achieving 89.52\%.

Table \ref{tab:acp_ener} summarizes the localized model predictions for energy consumption across the clients. The predicted and actual values are reported for each client, categorized into two emission levels (low and moderate).

\begin{table}[t]
\caption{Localized model predictions: Energy consumption}
\centering
\begin{small}
\begin{tabular}{ccc}
\toprule
\textbf{Client ID} & \textbf{Predicted Classes} & \textbf{Actual Classes} \\ \midrule
0               & {[}1, 0, 1, 1, 0{]}       & {[}1, 1, 1, 0, 0{]}    \\ 
1               & {[}0, 0, 0, 1, 1{]}       & {[}0, 0, 0, 1, 0{]}    \\ 
2               & {[}1, 1, 1, 0, 1{]}       & {[}1, 1, 0, 0, 1{]}    \\ 
3               & {[}0, 1, 0, 0, 1{]}       & {[}0, 1, 0, 0, 1{]}    \\ 
4               & {[}1, 0, 1, 1, 0{]}       & {[}1, 0, 1, 1, 1{]}    \\ 
5               & {[}0, 0, 1, 1, 0{]}       & {[}0, 0, 1, 0, 0{]}    \\ 
6               & {[}1, 1, 0, 1, 1{]}       & {[}1, 0, 1, 1, 0{]}    \\ 
7               & {[}0, 1, 1, 0, 0{]}       & {[}0, 1, 1, 0, 1{]}    \\ 
8               & {[}1, 0, 1, 1, 0{]}       & {[}1, 0, 1, 0, 0{]}    \\ 
\bottomrule
\end{tabular}
\end{small}
\vspace{0.2cm}
\label{tab:acp_ener}
\end{table}

The {Neural Network} consistently achieved the highest accuracy in both experiments. However, our proposed {N-tier encoded federated learning} method performed competitively and showed promise, particularly in decentralized settings with distributed datasets. 
Overall, the results demonstrate the effectiveness of our method in geo-spatial analysis and provide a basis for future refinements.

\section{Conclusion and Future Work}
\label{conclusion}

This research investigated the performance of an N-tier federated learning framework for predicting outcomes using spatial datasets across different regions. The results demonstrated varying predictive accuracy, with some regions performing better than others, highlighting the complexities involved in spatial data modeling.
The experiments with the gas emission dataset achieved 75.62\% accuracy, despite the uneven distribution of data, while the energy consumption dataset yielded a higher accuracy of 89.52\%, benefiting from a more evenly distributed dataset. These findings illustrate the potential of our approach but also underscore the need for improvements, especially when handling uneven data distributions.

This research contributes to the growing field of spatial prediction by showing the effectiveness of aggregated models across various regions. It underscores the crucial need for refining model architectures and improving data distribution to enhance predictive accuracy and robustness. This study highlights the significance of model validation in enabling more informed decision-making and policy development.

Several areas warrant further exploration for future work. First, integrating temporal dimensions into the spatial datasets would add depth to the predictions, enabling models to capture spatial and temporal variations. Additionally, incorporating more advanced features and refining neural network architectures could enhance predictive performance. Alternative techniques, such as ensemble or hybrid models, could also be explored to improve the capture of complex spatial patterns.
Moreover, addressing data heterogeneity and ensuring model generalization across diverse regions are critical to enhancing the reliability and scalability of the models. This could involve experimenting with different aggregation strategies, improving federated learning algorithms, or optimizing model parameters for better performance across all tiers.

\section*{Acknowledgment}
This work was supported by UNB University Research Fund (URF-NF-EXP 2024-16) and the Harrison McCain Young Scholars Award (HMF2023 YS-02).
\bibliographystyle{ieeetr}
\bibliography{references.bib}

\end{document}